\documentclass[conference, final]{IEEEtran}
\usepackage[noadjust]{cite}

\ifCLASSINFOpdf
  \usepackage[pdftex]{graphicx}
  \graphicspath{{figures/}}
  \DeclareGraphicsExtensions{.png,.pdf}
\else
\fi
\usepackage[cmex10]{amsmath}
\usepackage{array}

\usepackage[tight,footnotesize]{subfigure}
\usepackage{chngcntr}
\counterwithin{paragraph}{subsection}

\usepackage{changes}

\begin{document}
\title{Grounding object perception in a naive agent's sensorimotor experience}

\author{
\IEEEauthorblockN{Alban Laflaqui\`ere}
\IEEEauthorblockA{AI Lab, Aldebaran\\
43 rue du Colonel Avia, 75015 Paris, France\\
Email: alaflaquiere@aldebaran.com}
\and
\IEEEauthorblockN{Nikolas Hemion}
\IEEEauthorblockA{AI Lab, Aldebaran\\
43 rue du Colonel Avia, 75015 Paris, France\\
Email: nhemion@aldebaran.com}
}


\maketitle

\begin{abstract}
Artificial object perception usually relies on a priori defined models and feature extraction algorithms. We study how the concept of object can be grounded in the sensorimotor experience of a naive agent. Without any knowledge about itself or the world it is immersed in, the agent explores its sensorimotor space and identifies objects as consistent networks of sensorimotor transitions, independent from their context. A fundamental drive for prediction is assumed to explain the emergence of such networks from a developmental standpoint. An algorithm is proposed and tested to illustrate the approach.
\end{abstract}


%
\IEEEpeerreviewmaketitle

\section{Introduction}
The framework of developmental robotics addresses the difficult problem of how an agent can autonomously learn to interact with its environment and progressively acquire more and more complex skills. In such an attempt, the question of how the agent actually perceives its environment is fundamental but rarely addressed in the literature. Either the problem is bypassed, by providing the agent with a priori models and feature extraction algorithms (e.g. \cite{pal1993review,bucak2014multiple}), or avoided, by taking a behaviorist standpoint to study the system (\cite{barto1998reinforcement,nolfi2000evolutionary}).

The sensorimotor approach of perception proposes a re-definition of the nature of perception that naturally accounts for its grounding in sensorimotor experience and its acquisition by an autonomous agent (\cite{o2001sensorimotor,o2011red}). Although the theory is promising, its application is demanding. It requires an overhaul of what is commonly accepted in artificial perception and an intertwined analysis of phenomena that are usually considered independently (motor outputs and sensory inputs, modalities...). 

The sensorimotor re-formalization of different perceptive notions like sound localization \cite{bernard2012sensorimotor}, spatial perception \cite{laflaquiere2012non}, rigid displacements \cite{terekhov2013space}, color perception \cite{philipona2006color} or body schema \cite{laflaquiere2015learning} have already been proposed. Notably enough, those works focus on perceptive properties that are environment-independent: they relate to space, in which the environment is immersed, or to properties of the agent's structure and perceptive apparatus. In this paper, we propose to address the grounding of the concept of \emph{object} in an agent's sensorimotor experience. We believe it forms with the concept of \emph{space}, previously studied, a large part of an agent's perceptual experience: most of what we perceive are objects in space.
Contrarily to previous works, this paper directly focuses on characterizing properties of the environment. Instead of being limited to an abstract formalization of the concept, an algorithm will also be proposed to illustrate how it can pragmatically be applied in a system, opening the door to practical problems like object recognition.
Moreover, in line with the developmental approach, we'll explain how discovering objects in the environment can be the byproduct of satisfying a more fundamental drive for prediction, as proposed in (\cite{friston2010free, barsalou2008grounded}).

This work also relates to literature beyond the direct application of the sensorimotor approach of perception. A parallel can for instance be drawn with the scan-path theory proposed to account for visual perception \cite{noton1970theory}. In the same vein, previous works also attempted to take inspiration from saccadic human vision and to characterize objects based on spatially distributed local descriptors (\cite{paletta2005q, volpi2014active}).
\added{It is also related to developments in reinforcement learning focusing on the learning of sensorimotor predictors (\cite{sutton2011, sutton2012}) and on the generation of meta-states by cutting connections in a graph of transitions (\cite{mannor2004dynamic, kazemitabar2009automatic}). Note however that our goal lies beyond prediction, which is assumed to be a driving force, to focus on the sensorimotor structures that it can capture. The main objective of this paper is thus to identify which structure does correspond to the perceptive experience of "objects" by a naive agent. Moreover, a computational solution is proposed to illustrate how it can be captured by a naive agent.}

\added{The paper is divided in four main sections. First, a formalization of the problem at hand is introduced. Second a sensorimotor solution to the problem is proposed. Third, simulations of simple agent-environment systems are described to illustrate the application of the method. Fourth, the limits and future developments of the approach are discussed.}




\section{Problem formulation}
\label{sec:problemformulation}

\added{The goal of this paper is to explain how the concept of object can be fundamentally grounded in the sensorimotor experience of a naive agent. First, a formal definition of naive agent is introduced. Second, the concept of object is defined by looking at the constraints it imposes on the sensorimotor experience of the agent.}

\subsection{Naive agents}
\label{sec:naiveagents}

In order to limit as much as possible any bias that robot designers usually introduce in the perceptive system of their robot, the agents we consider have initially no a priori knowledge about themselves or the world. They are granted with the status of \emph{agent} thanks to their:
\begin{itemize}
\item motors, controlled by motor variables that form a motor configuration denoted $\mathbf{m}$, and that enable actions in the world,
\item sensors, whose all elementary excitations constitute a sensory state denoted $\mathbf{s}$, and that capture information about the environment,
\item data processing medium that supports the processing of sensorimotor data and potentially the following building of cognitive abilities.
\end{itemize}
A parallel can here be drawn with the minimal cognitive systems described by Beer \cite{beer2000dynamical}.

In the following, we assume that motors are controlled in position with no transient phase. Likewise, the sensory input is generated without transient phase. Moreover, we also assume that the mapping between the agent's motor space and the actual position of its sensors in space is homogeneous. This way, the same motor variation generates the same sensors' displacement, regardless of where it is performed in the motor space. Note that those limitations could be overcome by further developments of the method. For instance, a non-homogeneous motor-position mapping could be considered by initially learning rigid displacements of the agent \cite{terekhov2013space}.

Naive agents can't be complete \textit{tabula rasa}, as some intrinsic drives must be added to the system in order for it to engage in an interaction with the world \added{and to process the data it has access to}. Identifying such fundamental drives is a fundamental goal of developmental robotics. They are expected to allow the building of a progressively more and more complex cognitive system.
\added{In this work, the focus is put on the drive to process the data. The drive to interact with the world is thus assumed to be minimalist: random exploration. Note that more sophisticated policies could of course be considered \cite{oudeyer2007}. The choice of stochasticity however allows us to focus on the other drive which is expressed in a generic form: a drive to predict sensorimotor experiences.}
In other words, the agent tries to control as much as possible its \textit{Umwelt} which sums up to the sensory and motor data it has access to. A similar idea has been proposed in \cite{friston2010free} as a drive to reduce surprise. Regardless of the phrasing used, the outcome of this drive is an internal modeling of the environment in order to maximize predictability of sensorimotor experience.

\subsection{A sensorimotor definition of objects}
\label{sec:objectdefinition}

Defining what is an object for a naive agent and how it can be perceived is not a trivial problem. The conventional approach in robotics is to rely on models implemented on the robot by its designer. However, when such a priori information is lacking, a processing system has to be built in the agent starting from raw sensorimotor data.

In the seminal paper \cite{o2001sensorimotor} and latter developments, O'Regan proposed to redefine \emph{perception} as \emph{the mastering of sensorimotor contingencies}. In other words, perception arises when the agent knows how the sensory input would be transformed by its own actions.
The philosophical discussion raised by such a claim lies out of the scope of this paper. However, it implies some interesting properties for a naive agent.
First and foremost, sensorimotor contingencies can be learned and a naive agent can thus acquire the ability to perceive by exploring its environment (or more precisely its sensorimotor \textit{Umwelt}) and discovering such contingencies.
Second, sensorimotor contingencies are constrained by the physical reality of the world the agent lives in. Its perceptive experience thus derives from the way the agent can interact with the world, and not from properties of the sensory or data processing systems that support this interaction, as it is commonly assumed.
Finally, this innovative approach of perception fits naturally with the fundamental drive introduced in \ref{sec:naiveagents}. By discovering sensorimotor contingencies, the agent increases its ability to predict its future experience.

\begin{figure}[!t]
\centering
\includegraphics[width=0.45\linewidth]{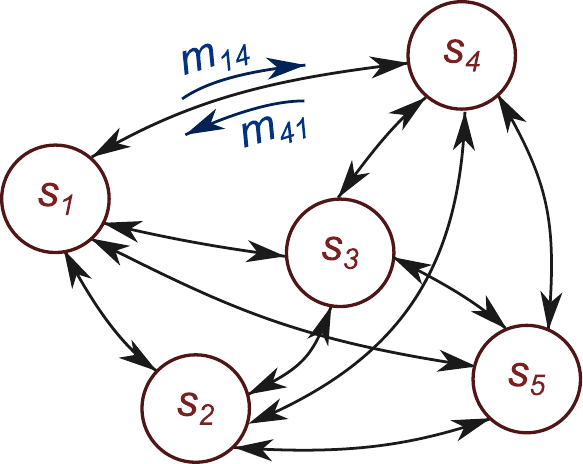}
\caption{Schematic representation of a sensorimotor network $O_k$ made of 5 sensory states and 10 highly probable motor transitions.}
\label{fig:SMgraph}
\end{figure}

Following such an approach, we propose to define an \emph{object} through the constraints it implies on the agent's sensorimotor interaction with the world. Let's assume as a first approximation that:
\begin{itemize}
\item an object is an extended cluster of matter with a rigid structure in the environment,
\item an object can be moved inside its environment or encountered in different environments.
\end{itemize}
For a naive agent exploring its sensorimotor space, this implies that it exists a subset of sensorimotor experiences whose structure is rigid and independent from the rest of the experiences. In other words, by identifying an object in the environment, the agent can accurately predict what sensory input it would get by doing given actions in a subpart of its sensorimotor space.
A higher-level analogy would be to say that by "seeing" the cap of a bottle, an agent is able to predict it would "see" its label by moving its visual sensor in a certain direction.
This is only true thanks to the very physical nature of an object and the way an agent can interact with it. Similarly to Poincar\'e saying that the notion of space would not emerge without the presence of rigid objects \cite{poincare1895espace}, we claim that the concept of object would not emerge without them having a rigid structure and being observed at different positions or in different environments.
This sensorimotor definition of an object can be formalized mathematically as:
\begin{eqnarray}
\label{eq:object}
&O_k = \big\{ \{\mathbf{s}_i\}, \{\Delta\mathbf{m}_{ij}\} \big\},\text{ with } i,j \in [1,I] \\
&\text{such that } P(\mathbf{s}_i, \Delta\mathbf{m}_{ij}, \mathbf{s}_j) \geq \alpha, \nonumber
\end{eqnarray}
where $O_k$ is the internal encoding of object $k$, $\{\mathbf{s}_i\}$ is a set of $I$ sensory states $\mathbf{s}_i$, \added{ $\{\Delta\mathbf{m}_{ij}\}$ is a set of $I^2$ motor transitions $\Delta\mathbf{m}_{ij}$ to go from $\mathbf{s}_i$ to $\mathbf{s}_j$,}
$P(\mathbf{s}_i,\Delta\mathbf{m}_{ij},\mathbf{s}_j)$ is the probability of experiencing $\mathbf{s}_j$ when starting from $\mathbf{s}_i$ and performing the motor transition $\Delta\mathbf{m}_{ij}$, and $\alpha$ is a high probability threshold.
A graphical illustration of a set $O_k$ is proposed in Fig.~\ref{fig:SMgraph}.

Note that the value $\alpha$ is purposely fuzzily defined. First, the probabilistic formulation of the transition allows the definition of confidence in the set and its potential reinforcement. Second, the value of $\alpha$ is system dependent and should be set accordingly to the maximal certainty an agent can achieve in predicting its sensorimotor experience. For instance, its value should be high for a noise-free agent-environment system but low if the signal-noise ratio is poor. Beyond this pragmatic consideration, the goal of Eq.~\ref{eq:object} is to capture the fact that interacting with an object implies regularities and thus predictability in the way the agent can transition between sensorimotor states. Given the fundamental drive for prediction introduced in \ref{sec:naiveagents}, such a phenomenon is valuable for the agent.

\section{Discovering objects}
\label{sec:discoveringobjects}

In order to illustrate the sensorimotor approach of objects promoted in this paper, we propose a proof-of-concept algorithm allowing a naive agent to discover highly probable sets of sensorimotor transitions in the experience it gathers while exploring the world.

Beforehand, let's introduce the assumption on which the method is based: changes in the world (object(s) moving or environment changing) have a low probability compared to changes in the agent's motor configuration. In other words, the agent has enough time to explore its sensorimotor space before any new change occurs  in the world \footnote{This constraint could be relaxed to ensure that the agent has time to at least explore a significant subpart of its sensorimotor space between two changes in the world.}. This hypothesis seems realistic for an agent that explores its environment and discovers objects, like a baby does while lying in his bed and playing with toys. The rest of its environment can indeed be considered static while the toys are moved or while he moves himself. A contrario, we claim that it would be impossible to discover the concept of object if the whole world was constantly changing at a fast rate.

\subsection{The algorithm}
\label{sec:algorithm}

\paragraph{Initialization}
The naive agent is placed in an environment where at least one object is present.
It explores the scene, formed by the environment and the object(s) it contains, by changing its motor configuration $\mathbf{m}$. Each motor configuration $\mathbf{m}_i$ explored this way is associated with a sensory state $\mathbf{s}_i$. As the predictability of sensorimotor transitions is the focus of this work, this whole set of experiences is stored as triplets $\{\mathbf{s}_i,\Delta\mathbf{m}_{ij},\mathbf{s}_j\}$ in the agent's memory.

As such, the agent explored only a single scene which can consequently be considered as a single object. Indeed it forms a sensorimotor network where each transition is highly predictable. In order to discover that it is not the case, it has to experience changes in the scene.

\paragraph{Experiencing changes}
After the scene has been completely explored, a change in the environment is generated. It consists in either moving the object(s) to a different position, or changing the environment that contains the object, or both. Such a change is noticeable by the agent as it implies a difference between new sensorimotor experiences and the knowledge it stored in memory. 

Similarly to what it did previously, it explores the scene to assess the existence of new sensorimotor transitions and potentially reinforce the ones it already stored. Every time an incoming sensory input $\mathbf{s}_i$ corresponds to one it already has in memory, it actively checks the relative transitions $\{\mathbf{s}_i,\Delta\mathbf{m}_{ij},\mathbf{s}_j\}$ in its memory. It does so by performing the motor command $\Delta\mathbf{m}_{ij}$ and verifying if the new incoming sensory state is $\mathbf{s}_j$.
If it is the case, the probability of the transition $\{\mathbf{s}_i,\Delta\mathbf{m}_{ij},\mathbf{s}_j\}$ is increased, while it is decreased otherwise. In the following, the probability of each transition $P(\mathbf{s}_i,\Delta\mathbf{m}_{ij},\mathbf{s}_j)$ is computed as the number of scenes in which the transition was valid divided by the number of scenes explored.
The same overall process is reproduced for multiple changes of the environment.

\paragraph{Identifying sensorimotor subsets}
\added{
All the probabilities can be stored in a 3D matrix $D$ with entries $\mathbf{s}_i,\mathbf{s}_j,\Delta\mathbf{m}_{ij}$ where $\mathbf{s}_i$ corresponds to a starting sensory state, $\mathbf{s}_j$ to a final sensory state, $\Delta\mathbf{m}_{ij}$ to a motor command, and each value corresponds to the probability of the given transition $P(\mathbf{s}_i,\Delta\mathbf{m}_{ij},\mathbf{s}_j)$. Note that the knowledge about predictable sensorimotor transitions is intrinsically contained in this matrix. It would thus be a satisfying result for the agent. However, an additional processing is performed for us to visualize the algorithm's outcome.}

\begin{figure*}[!t]
\centering
\includegraphics[width=0.85\linewidth]{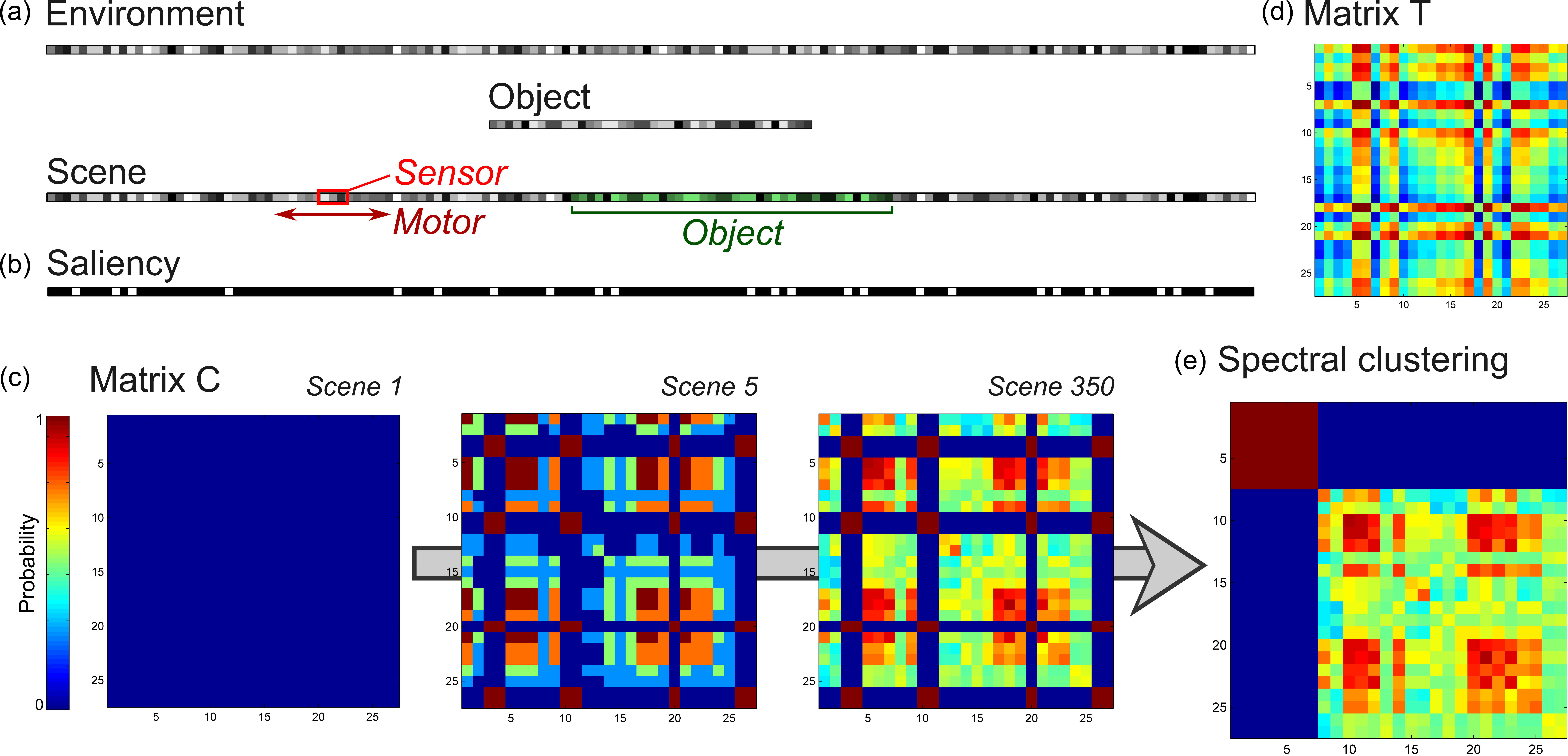}
\caption{Simulation 1. (a) \added{Generation of a random scene }and agent's exploratory capacities. The object is colored only to simplify visualization. (b) Map of positions for which the sensory input is salient. (c) Evolution of matrix $C$ of probabilities of transitions during exploration of successive scenes. \added{The matrix $C$ is initially built based on salient inputs found in the first scene explored.} (d) Motor transition matrix. (e) Reordering of matrix $C$ based on its spectral clustering.}
\label{fig:1Denvironment}
\end{figure*}

\subsection{Results visualization}
\label{sec:resultsvisualization}
\added{The final step of the algorithm is to identify in the matrix $D$ the sets of sensory states interconnected with high probable transitions and visualize them. Due to the way it is built, the matrix $D$ is such that only one probability $P(\mathbf{s}_i,\Delta\mathbf{m}_{ij},\mathbf{s}_j)$ is non-zero for each pair of entries $\{\mathbf{s}_i,\mathbf{s}_j\}$. In order to simplify the analysis, the matrix $D$ is thus reduced into the 2D matrix $C$ by dropping the third dimension and keeping only the non-zero probability for each entries $\{\mathbf{s}_i,\mathbf{s}_j\}$. The removed motor components are stored in a matrix $T$ where each row corresponds to a starting sensory input $\mathbf{s}_i$, each column corresponds to a final sensory state $\mathbf{s}_j$, and each value corresponds to the relative motor transition $\Delta\mathbf{m}_{ij}$. Intuitively, this simplification of matrix $D$ into $C$ means that we only need to know that sensory states are linked with given probabilities to determine interconnected sets of sensory inputs; the actual motor commands values are irrelevant for this analysis.
}

A spectral clustering of the matrix $C$ is performed to assess the existence of highly probable sets of interconnected transitions. The method determines in a similarity matrix clusters with high intra-connectivity and low extra-connectivity \cite{ng2002spectral}. We thus suppose that a highly probable transition represents a high similarity between the corresponding sensory states and inversely.
The spectral clustering projects the data in a space of dimension $k$ in which it can be clustered by any unsupervised clustering method, commonly a simple $k$-means. The value $k$ is set accordingly to the number of expected clusters. Such a hand-tuned parameter is not suitable for a fully autonomous agent. However, spectral clustering is mainly used here for visualization purpose. If it was a core component of the data processing, an autonomous way to determine the dimension $k$ would be to search for the number of significant eigenvalues of the similarity matrix $C$ \cite{lee2007nonlinear}.

Finally, the rows and columns of the matrix $C$ are reordered by gathering the ones belonging to the same spectral cluster. Any set of highly probable interconnected transitions should thus appear as a full square of high values on the diagonal of the reordered matrix $C$. Although such a reordering processing is not required for the agent -- for which the knowledge of the transitions probabilities is sufficient -- it allows an easier visualization of the exploration's outcome.

In the following, simulations are presented to illustrate the application of the algorithm and how the presence of object(s) in the environment can be captured by a naive agent.

\section{Experiments}
\label{sec:experiments}

Two experiments are presented hereafter to test the algorithm proposed in \ref{sec:algorithm}. For each of them, the simulated agent-environment system is presented, the exploratory process is run and the results of the spectral clustering are discussed.

\subsection{Simulation 1: 1D world}
\label{sec:Simulation1}

The first experiment proposes a simple system and scenario in order to focus on the core idea of the algorithm. 

\paragraph{The system} The world in which the agent and the environment live is a one dimensional space made up of $150$ successive elements. Each element can have physical properties represented by a single value $x$ in a finite set of $10$ integers from $1$ to $10$. This could for instance correspond to luminosity if the agent has a photo-sensible sensor, or different textures in the case of a tactile sensor.
Note that a discrete description of the world is proposed here for the sake of simplicity but a continuous one would also be compatible with the approach.

In this world, an environment is defined as a state of the world where each element's value is randomly drawn. An object made from $40$ adjacent elements with fixed random values is placed in the world, overlapping the ones from the environment and forming a "scene" that the agent can explore.  
The agent's sensor is made of three adjacent cells sensible to the values of elements in the world and generates the sensory state $\mathbf{s} = [x_1,x_2,x_3]$. The position of this sensor is set by the agent's motor. Its configuration is denoted $\mathbf{m} = m_1$, with $m_1$ the single motor variable controlled by the agent. Given the discrete nature of the world, the sensor is supposed to move with discrete steps the length of one or multiple elements. 
An illustration of the different components of the simulation is presented in Fig.~\ref{fig:1Denvironment}(a).

In order to limit the computational cost of the algorithm application, we assume that some saliency of the sensory input can be defined. This way, only salient sensory inputs need to be taken into account during the processing, limiting this way the size of the matrix $C$. The saliency definition could stem from multiple considerations: low-level statistical properties of the input, top-down influence on the input relevance, optimization of some criterion on an evolutionary timescale. This discussion however lies out of the scope of this paper. We thus simply define an arbitrary criteria and remind the reader that its only purpose is to limit the computational cost of the simulation. A sensory state is considered salient and processed by the agent if its convolution with the contrast filter $[-0.5, 1, -0.5]$ is greater than $0.4$. A mapping of salient sensory inputs is illustrated in Fig.~\ref{fig:1Denvironment}(b) for a given scene.

\begin{figure}[t!]
\centering
\includegraphics[width=0.5\linewidth]{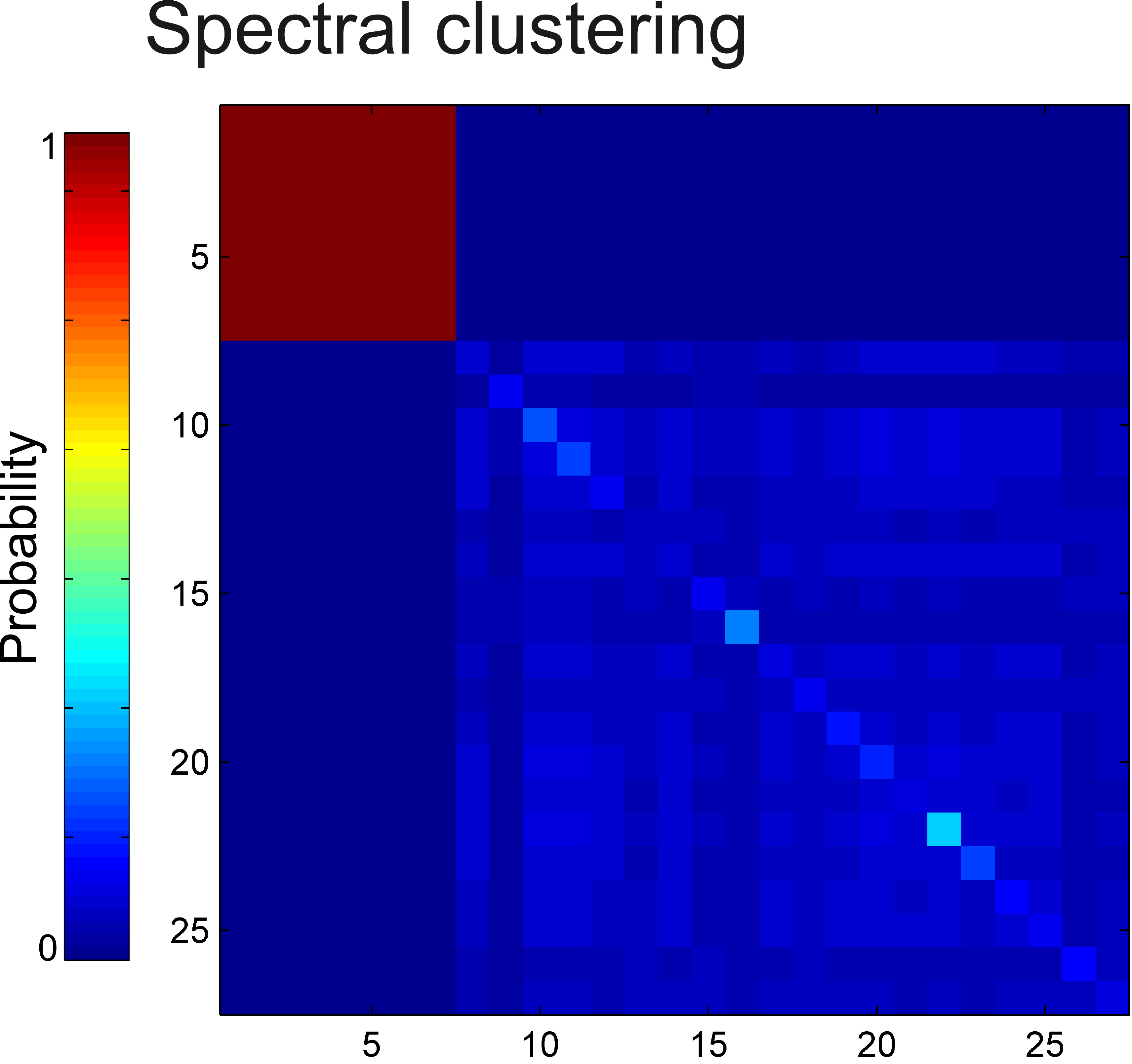}
\caption{Resulting reordered matric $C$ with a $5\%$ probability for the environment to change between two scenes.}
\label{fig:1DChangingBkg}
\end{figure}

\paragraph{The exploration} The exploratory scenario proposed in \ref{sec:algorithm} is performed.
The exploration of the first scene leads to the creation of an all-ones matrix $C$ \added{of size $27\times 27$, $27$ being the number of salient inputs discovered in the scene}. \added{The matrix $T$ representing the sensorimotor transitions $\{\mathbf{s}_i,\Delta\mathbf{m}_{ij},\mathbf{s}_j\}$ is presented in Fig.~\ref{fig:1Denvironment}(d)}.

The agent explores every new scene that it encounters after a change has been generated by randomly changing the object's position in the world. The evolution of matrix $C$ is presented in Fig.~\ref{fig:1Denvironment}(c).
\added{We can see that between scene 1 and 5, the matrix $C$ has already changed: some probabilities of transition are high (red), as they have been experienced in all scenes so far, while others are low (blue) as the agent has not been able to experience them in each scene.}
After $350$ changes, exploration is stopped and the data collected by the agent is analyzed.

\paragraph{Results} The reordered matrix $C$ obtained after spectral clustering is presented in Fig.~\ref{fig:1Denvironment}(e). Two clusters appear clearly.
The first one displays a maximal probability of all intra-transitions and corresponds to the object. This last is thus internally represented by a network of $7$ sensory states and the corresponding motor transitions between them.
The second cluster corresponds to the environment. Althought the goal of the algorithm is to capture the existence of objects in the world, the environment has been internally represented as one due to its staticity. Indeed, it stayed unchanged during the whole exploration. The probability of transition between the sensory inputs initially discovered in the environment thus remained significantly high. They are not as high as for the object though, due to the fact that the object can randomly hide some salient points in the background.

The same simulation has been run adding a probability of $5\%$ for the environment to randomly change in every new scene. The resulting clustering is presented in Fig.~\ref{fig:1DChangingBkg}. The internal representation of the object appears unchanged but the cluster corresponding to the environment now contains low probabilities. This is due to the low probability of experiencing the same sensorimotor transitions in different environments.

\added{Some additional remarks on the results are noteworthy. First, note that the number of salient sensory inputs in the network associated with an object is only an indirect characterization of its size. Indeed, the larger the object, the greater the probability that it contains salient inputs but one could easily imagine large objects with only few salient inputs. Second, changing the saliency criteria would modify the number (and nature) of the sensory inputs stored in the agent's memory. However, the outcome of the algorithm would still be qualitatively the same: a (different) cluster of sensory inputs would appear in the transition matrix as an encoding of the object.
Finally, the hypothesis of complete exploration for each scene introduced in section \ref{sec:discoveringobjects} could be loosened without affecting too significantly the algorithm's outcome. Each transition would have a probability of not being explored in each scene, which means that the maximum possible value in the matrix $C$ would be statistically lower. However, the network of transitions related to the object would still have a greater probability than other transitions to or in the background.
}

\subsection{Simulation 2: 2D world with multiple objects }
\label{sec:Simulation2}

The algorithm proposed in \ref{sec:algorithm} has been formulated in a generic way. In this second simulation, we introduce a more complex system to illustrate how it extends to more degrees of freedom and multiple objects in the world.

\begin{figure*}[!t]
\centering
\includegraphics[width=0.9\linewidth]{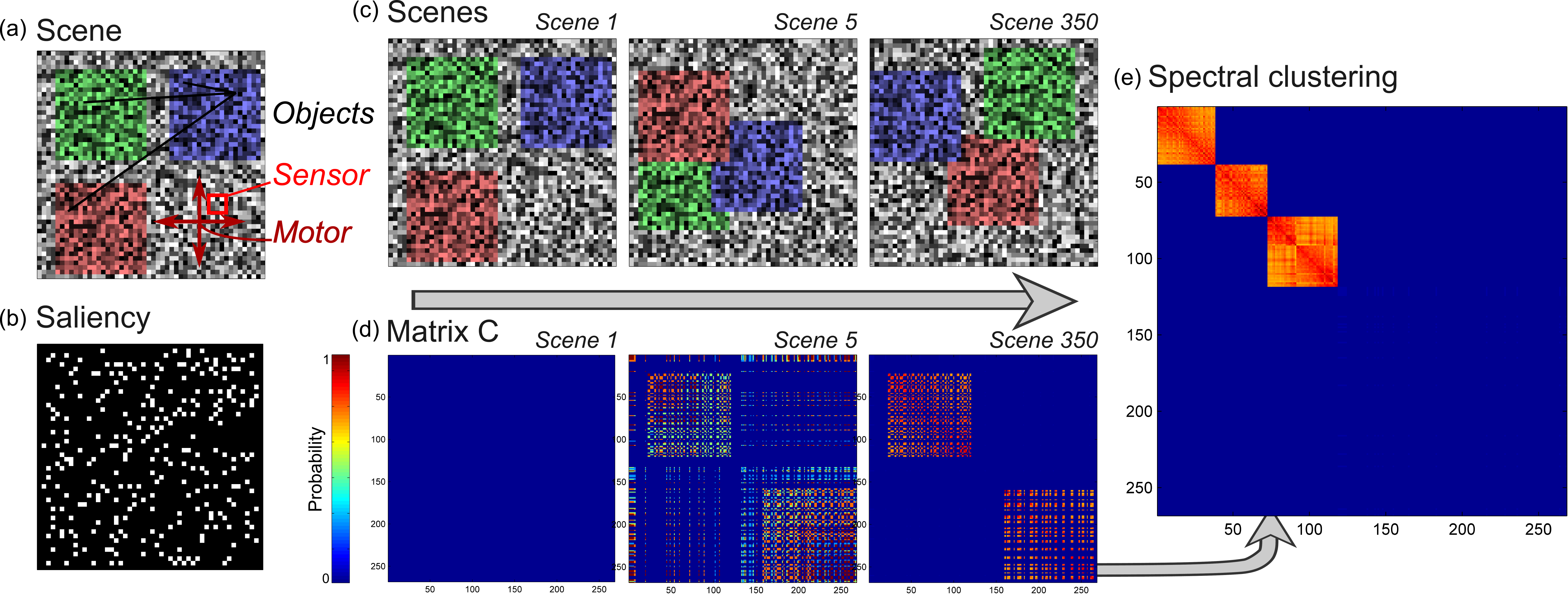}
\caption{Simulation 2. (a) Scene generation and agent's exploratory capacities. Objects are color to simplify visualization. (b) Positions of salient sensory inputs. (c) Evolution of the scene during simulation. (d) Evolution of matrix $C$ during exploration of the successive scenes. (e) Reordering of matrix $C$ based on its spectral clustering.}
\label{fig:2D3Objects}
\end{figure*}

\paragraph{The system} The world is similar to the one described in simulation $1$ but extended to a two dimensional space where $50\times50$ square elements are arranged in a grid. Three different square objects are placed in the world. Each of them is made up of $20\times20$ elements with random values. This constitutes a more realistic scenario as environments in the real world are made up of multiple independent objects. The objects can overlap in a random order that is drawn every time they move. Note that objects could be of any shape, although squares have been considered for simplicity.

The agent's sensor is extended to a $3\times3$ grid of cells that generates the sensory input $\mathbf{s} = [x_1,\dots,x_9]$. A new arbitrary filter is introduced to determine the saliency of any sensory state: 
\begin{equation}
\left[
\begin{array}{ccc}
-1/16 & -3/16 &-1/16\\
-3/16 & 1 & -3/16\\
-1/16 & -3/16 & -1/16
\end{array}
\right], \nonumber
\end{equation}
The sensors position is controlled by two motor variables that form the motor configuration $\mathbf{m} = [m_1,m_2]$.
An illustration of the different components of the simulation is presented in Fig.~\ref{fig:2D3Objects}(a,b). The overall system can be interpreted as a very basic visual system where an eye would be saccading on a flat visual scene.

\paragraph{The exploration}
The exploratory scenario proposed in \ref{sec:algorithm} is performed identically to the first simulation. A single constraint is added to ensure that the objects don't overlap in the initial scene in order to guarantee that they are completely explored and that no relevant sensorimotor transition would be missing for further reinforcement.

A series of $350$ scene changes are performed. Each change is generated by randomly moving the objects and by changing the environment with a probability of $5\%$. The evolution of the scenes the agent can explore along with the evolution of the matrix $C$ are represented in Fig.~\ref{fig:2D3Objects}(c,d).

\paragraph{Results}
The reordered matrix $C$ obtained after spectral clustering is presented in Fig.~\ref{fig:2D3Objects}(e). Three clusters of highly probable intra-transitions appear clearly as internal representations of the three objects. Contrarily to simulation 1, those probabilities are not necessarily maximal as objects can eventually overlap and hide relevant sensory inputs in some scenes.
As in Fig.~\ref{fig:1DChangingBkg}, a last cluster corresponds to the environment but with too low intra-transitions probabilities to appear clearly.

\section{Conclusion}
\label{sec:conclusion}

In this paper, we proposed a definition of the concept of object in the sensorimotor framework of perception. Based on the constraints it imposes on an agent's interaction with the world, an object can be described as a network of highly probable sensorimotor transitions. Moreover, we explained how such an internal construction can occur in a naive agent. Assuming a fundamental drive for prediction, the concept of object is the fortunate byproduct of the agent's attempt to control its sensorimotor experience.

In order to illustrate this new approach, we proposed a proof-of-concept algorithm to discover networks of highly probable sensorimotor transitions. The algorithm has been tested on two simple simulated systems where the agent can interact with objects in different contexts (positions of the objects or environments that contain them). While interacting, the agent tries to extract subsets of its sensorimotor experience where predictability is high. In the current algorithm's outcome, they can be visualized as clusters along the diagonal of a probability matrix of transitions.

\added{According to the sensorimotor contingencies theory, perception doesn't derive from the properties of the sensory apparatus but from the properties of the agent's interaction with the world. In the current work, this means that the "object" definition is relevant regardless of the actual hardware properties of the agent. The use of different sensors and motor would also lead to the building of highly probable networks of transitions. Although certainly different, it would be a relevant encoding of the corresponding objects from the agent's internal point of view. For instance, the definition of object proposed in this paper is viable with either a visual sensor  or a tactile one.}
\added{Moreover, although the discovery of objects naturally leads to their dissociation from the background, the process proposed in this work is notably different from traditional object segmentation approaches. While the last rely on cutting the object out of the scene \cite{cremers2007}, the sensorimotor approach on the contrary identifies consistent relations in a subpart of the scene. Fundamentally, this approach thus progressively identify structures in the sensorimotor experience, which is otherwise un-interpretable by a naive agent. This way, it aims at building an intrinsically grounded semantics of the world.
The sensorimotor approach also differs from other methods designed to learned sensorimotor predictors (for instance \cite{sutton2012}). Indeed, it goes beyond the estimation of predictors, whose actual implementation could be done in very different ways, to focus on identifying the structure underlying those predictions. It is this structure that characterizes the agent's ability to interact with the world, and thus to perceive it.}

Many future developments are considered to improve the current algorithm.
First and foremost, it should be extended to allow the continuous acquisition of new sensorimotor transitions in the agent's memory as it explores the world. This would remove the artificial asymmetry between the exploration of the first scene, which is memorized, and the consecutive scenes, which are only explored to reinforce this memory. However, it would also raise new difficulties. In particular, the same sensory state can be present multiple times in a single object and/or in multiple environments and objects. The agent should thus determine to which network it belongs when processing the data. A solution to this problem is to not look at single sensorimotor transitions but more global sensorimotor \emph{contexts}, \added{which are sets of co-occuring sensorimotor experiences}. A re-formalization of the approach in terms of dynamical systems would also be beneficial to deal with this continuous growth of the memory and the temporally extended nature of sensorimotor contexts \cite{buhrmann2013dynamical}. Such a dynamical framework would also more easily manage a stream of continuous sensory data, making the saliency processing introduced in this paper unnecessary.

Second, the notion of sensory state has to be refined as only a subpart of the sensory input $\mathbf{s}$ might be relevant to characterize the object explored. It is the case for instance in the second simulation (see \ref{sec:Simulation2}) when the sensor is centered on an object's corner (only $4$ over $9$ sensory cells capture properties of the objects).
\added{Variability of the sensory inputs corresponding to one feature of the object (due to noise, lighting condition...) also needs to be taken into account in the definition of sensory state. This additional flexibility would be the main step towards an application on a real robotic platform.}

Finally, the sensorimotor network, which is already translation invariant, should be made rotation invariant. The way to solve this problem is to have the agent discover the set of displacements (translations and rotations) it can perform in space, as proposed in \cite{laflaquiere2012non,terekhov2013space}.
\added{This mastering can lead the agent to discover that the sensorimotor network defining an object is invariant to those displacements. This potentially includes displacements of objects in 3D with the additional constraint that some subparts of the networks corresponding to non-visible faces of the objects are not necessarily accessible to the agent.}
Such a bridge between the problems of space and object perception would also emphasize their intertwined nature and shed a new light on the way a naive agent can acquire the ability to perceive its environment.

\bibliographystyle{IEEEtran}
\bibliography{ICDL2015}

\end{document}